\documentclass[letterpaper, 10 pt, journal, twoside]{IEEEtran}

\usepackage{times}

\usepackage{amsmath,amsfonts,bm}

\def\eqref#1{equation~\ref{#1}}

\def\1{\bm{1}}

\DeclareMathAlphabet{\mathsfit}{\encodingdefault}{\sfdefault}{m}{sl}
\SetMathAlphabet{\mathsfit}{bold}{\encodingdefault}{\sfdefault}{bx}{n}

\usepackage{graphicx}
\usepackage{booktabs}
\usepackage{subcaption}

\usepackage{url}
\usepackage{float}
\usepackage{wrapfig}
\usepackage{footnote}

\usepackage{color}
\usepackage{xcolor}
\usepackage{cite}
\definecolor{citecolor}{HTML}{094AA8}
\newcommand{\revise}[1]{{\color{black}{ #1}}}
\newcommand{\blacklink}[2]{\href{#1}{\color{black}{#2}}}

\usepackage[pagebackref=true,breaklinks=true,letterpaper=true,urlcolor=citecolor,colorlinks,citecolor=citecolor,bookmarks=false,linkcolor=citecolor]{hyperref}

\renewcommand{\thefootnote}{\alph{footnote}}
\newcommand{\astfootnote}[1]{
\let\oldthefootnote=\thefootnote
\setcounter{footnote}{0}
\renewcommand{\thefootnote}{\fnsymbol{footnote}}
\footnote{#1}
\let\thefootnote=\oldthefootnote
}

\newcommand\blfootnote[1]{
  \begingroup
  \renewcommand\thefootnote{}\footnote{#1}
  \addtocounter{footnote}{-1}
  \endgroup
}

\begin{document}

\title{Visual Reinforcement Learning with Self-Supervised 3D Representations}

\author{Yanjie Ze$^{*1}$ \quad Nicklas Hansen$^{*2}$ \quad Yinbo Chen$^2$ \quad
Mohit Jain$^2$ \quad  Xiaolong Wang$^2$}

\markboth{IEEE Robotics and Automation Letters. Preprint Version. Accepted March, 2023}
{Ze \MakeLowercase{\textit{et al.}}: Visual Reinforcement Learning with Self-Supervised 3D Representations}

\twocolumn[{
\renewcommand\twocolumn[1][]{#1}
\maketitle
\vspace{-0.1in}
\begin{center}
\begin{minipage}{0.685\textwidth}
        \centering
        \begin{minipage}{0.025\textwidth}
            \centering
            \rotatebox{90}{
\small 
                \centering
                Real (Perturb)~~~~~~~~~Real~~~~~~~~~~~~~Sim~~~~~~~~~~~~~}
            
        \end{minipage}\hspace{0.01in}
        \begin{minipage}{0.23\textwidth}
            \centering
            \textbf{Reach}\vspace{0.03in}\\
            \includegraphics[ width=\textwidth]{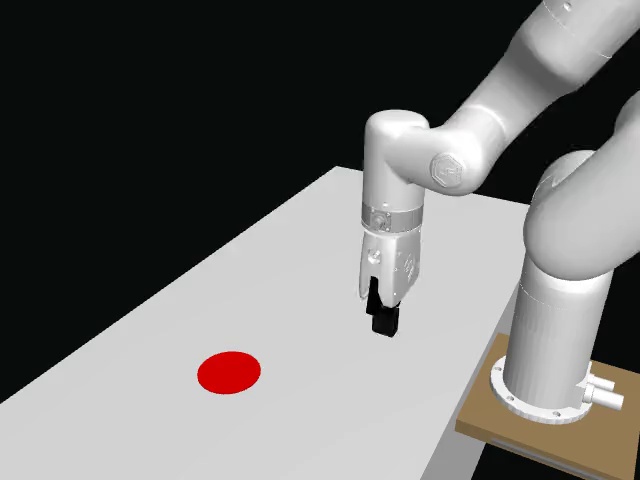}\vspace{0.005in}\\
         
            \includegraphics[ width=\textwidth]{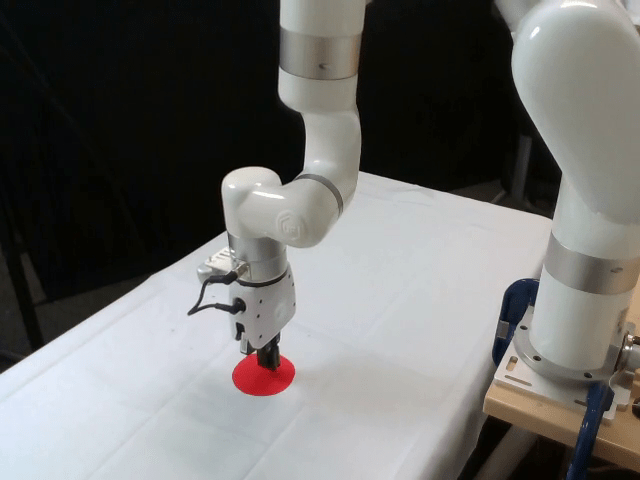}\vspace{0.01in}\\
            \includegraphics[width=\textwidth]{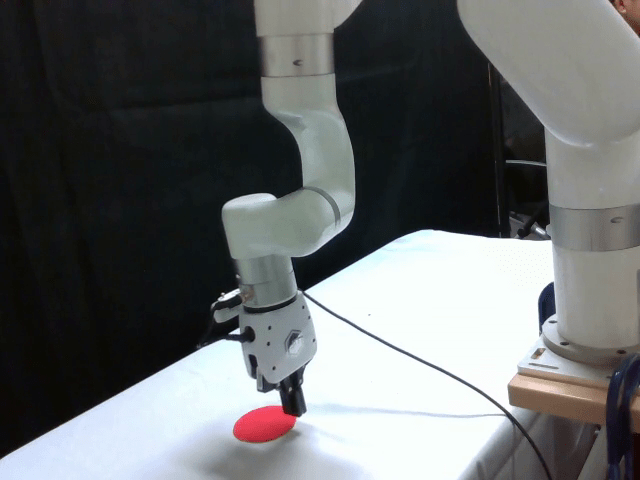}
        \end{minipage}\hspace{0.01in}
        \begin{minipage}{0.23\textwidth}
            \centering
            \textbf{Push}\vspace{0.03in}\\
            \includegraphics[width=\textwidth]{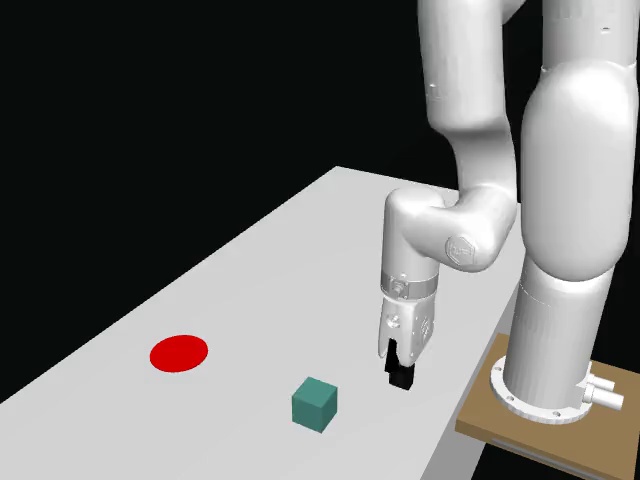}\vspace{0.005in}\\
           
            \includegraphics[ width=\textwidth]{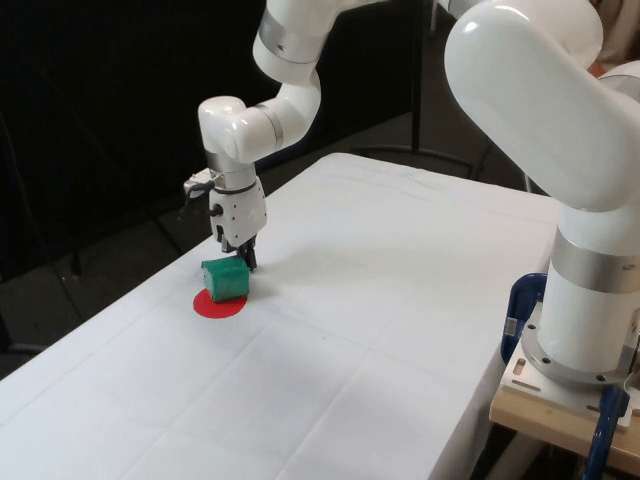}\vspace{0.01in}\\
            \includegraphics[width=\textwidth]{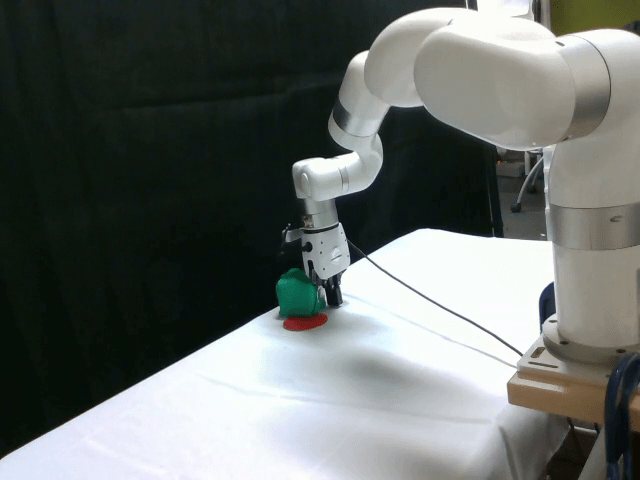}
        \end{minipage}\hspace{0.01in}
        \begin{minipage}{0.23\textwidth}
            \centering
            \textbf{Peg in Box}\vspace{0.005in}\\
            \includegraphics[ width=\textwidth]{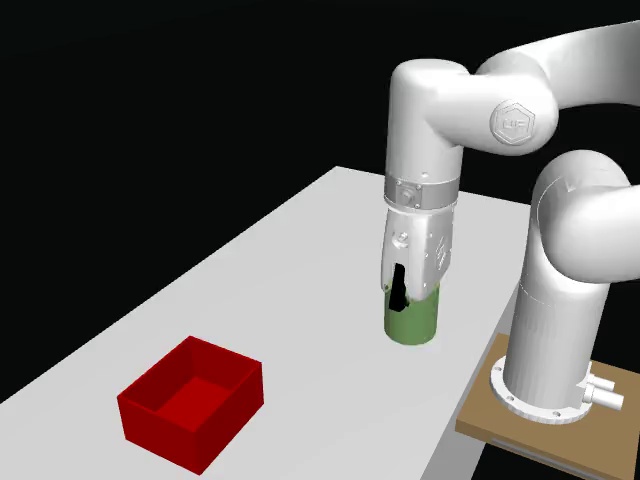}\vspace{0.005in}\\
           
            \includegraphics[width=\textwidth]{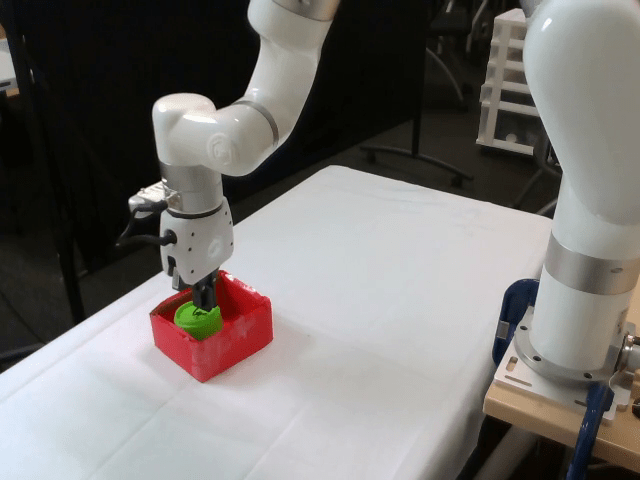}\vspace{0.01in}\\
            \includegraphics[width=\textwidth]{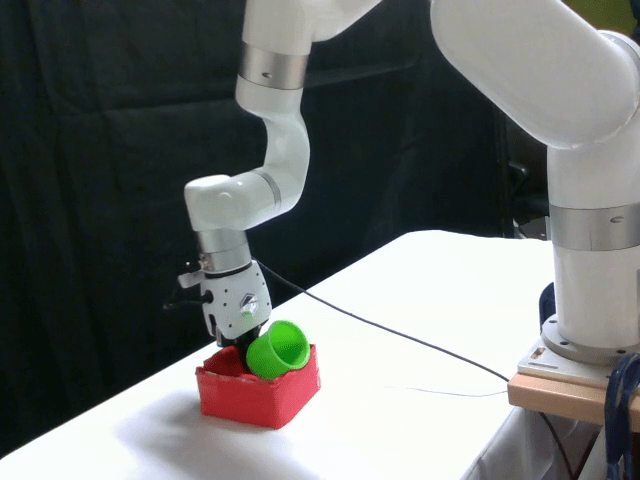}
        \end{minipage}\hspace{0.01in}
        \begin{minipage}{0.23\textwidth}
            \centering
            \textbf{Lift}\vspace{0.03in}\\
            \includegraphics[width=\textwidth]{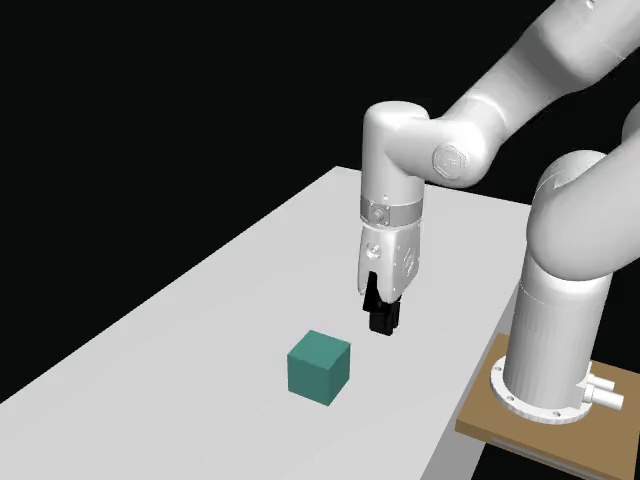}\vspace{0.005in}\\
            \includegraphics[width=\textwidth]{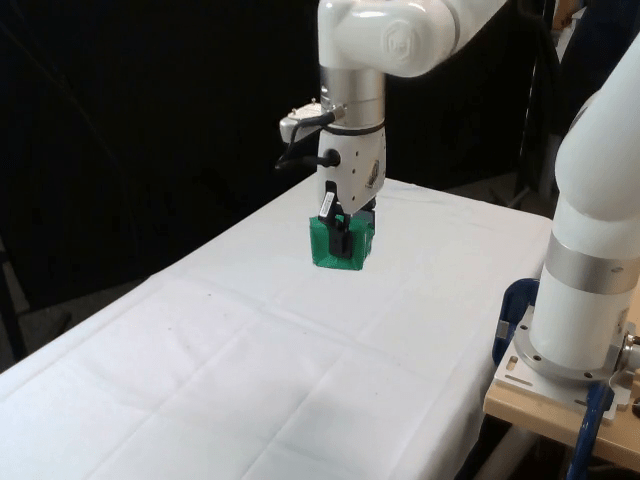}\vspace{0.01in}\\
            \includegraphics[width=\textwidth]{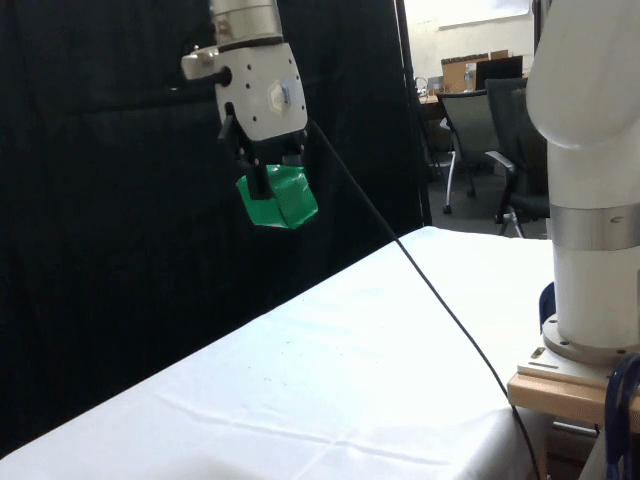}
        \end{minipage}
    \end{minipage}\hspace{0.05in}
    \begin{minipage}{0.2925\textwidth}
        \centering
        \textbf{Robot setup}\vspace{0.005in}\\
        \includegraphics[ width=\textwidth]{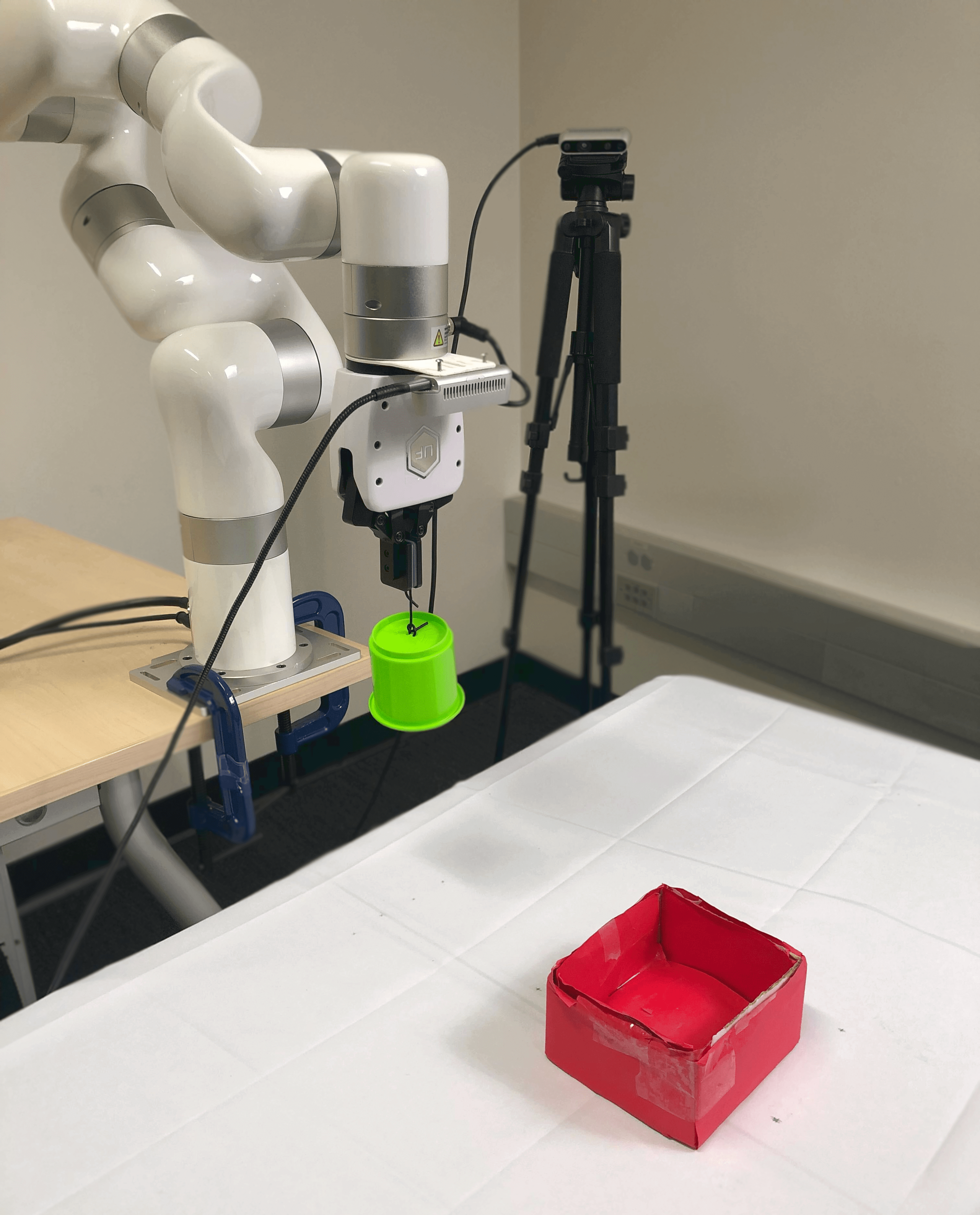}
    \end{minipage}
    \captionof{figure}{\textbf{Overview of sim-to-real tasks.} We consider four tasks for our sim-to-real experiments: \emph{(1) reach}, \emph{(2) push},  \emph{(3) peg in box},  and \emph{(4) lift}. Observations are captured by a static over-the-shoulder camera (pictured). We visualize the initial configuration of robot and objects in simulation and the success in the real world. \revise{We consider transfer to two distinct real-world setups with varying degrees of similarity to the simulation, namely in terms of camera view and lighting.}
  }
    \label{fig:sim2real}
    \vspace{0.1in}
\end{center}
}]

\begin{abstract}
A prominent approach to visual Reinforcement Learning (RL) is to learn an internal state representation using self-supervised methods, which has the potential benefit of improved sample-efficiency and generalization through additional learning signal and inductive biases. However, while the real world is inherently 3D, prior efforts have largely been focused on leveraging 2D computer vision techniques as auxiliary self-supervision. In this work, we present a unified framework for self-supervised learning of 3D representations for motor control. Our proposed framework consists of two phases: a \textit{pretraining} phase where a deep voxel-based 3D autoencoder is pretrained on a large object-centric dataset, and a \textit{finetuning} phase where the representation is jointly finetuned together with RL on in-domain data. We empirically show that our method enjoys improved sample efficiency compared to 2D representation learning methods. Additionally, our learned policies transfer zero-shot to a real robot setup with only approximate geometric correspondence, and successfully solve motor control tasks that involve grasping and lifting from \textit{a single, uncalibrated RGB camera}. Code and videos are available at {\small\url{https://yanjieze.com/3d4rl/}}.

\blfootnote{Manuscript received: December, 04, 2022; Revised February, 26, 2023; Accepted March, 03, 2023.\\
\indent $^{*}$Equal contribution. $^1$Shanghai Jiao Tong University, Shanghai, China, work done while an intern at University of California San Diego. $^2$University of California San Diego, CA, USA.  Correspondence at \blacklink{emailto:xiw012@ucsd.edu}{\texttt{xiw012@ucsd.edu}}.\\ 
\indent This paper was recommended for publication by Editor Jens Kober upon evaluation of the Associate Editor and Reviewers' comments.\\
\indent  This work was supported, in part, by Amazon Research Award and gifts from Qualcomm. \\
\indent Digital Object Identifier (DOI): see top of this page.

}
\end{abstract}
\vspace{-0.1in}

\IEEEpeerreviewmaketitle

\vspace{-0.05in}
\section{Introduction}

\label{sec:introduction}

While deep Reinforcement Learning (RL) has proven to be a powerful framework for complex and high-dimensional control problems, most notable successes have been in problem settings either with access to fully observable states \cite{Lillicrap2016ContinuousCW, silver2017mastering, Andrychowicz2020LearningDI}, or settings where partial observability through 2D image observations (\emph{visual} RL) suffice, e.g., playing video games \cite{mnih2013playing}. While potential applications of visual RL are far broader, it has historically been challenging to deploy in areas such as robotics, in part due to the complexity of controlling from high-dimensional observations.

A prominent approach is to tackle the resulting complexity by learning a good representation of the world, which reduces the information gap that stems from partial observability. Leveraging techniques such as self-supervised objectives for joint representation learning together with RL has been found to improve both sample efficiency \cite{yarats2019improving, laskin2020curl} and generalization \cite{Higgins2017DARLAIZ, Nair2018VisualRL, hansen2021softda} of RL in control tasks. Recently, researchers also discover training RL from embeddings produced by pretrained frozen visual encoders trained on external datasets can match the performance of \emph{tabula rasa} (from scratch) representations while requiring less in-domain data \cite{Xiao2022MVP, Parisi2022TheUE}.

Yet, efforts have largely been focused on applying successful techniques from 2D computer vision to control problems. However, our world is inherently 3D and agents will arguably need to perceive it as such in order to tackle the enormous complexity of real world environments \cite{dobbins1998distance, cheng2018reinforcement,akinola2020learning}. For example, a robot manipulating objects may encounter challenges such as partial occlusion and geometric shape understanding, neither of which are easily captured by 2D images without prior knowledge or strong inductive biases \cite{wang20206, tung2019learning}.

\begin{figure*}[t]
    \centering
    \includegraphics[width=0.9\textwidth]{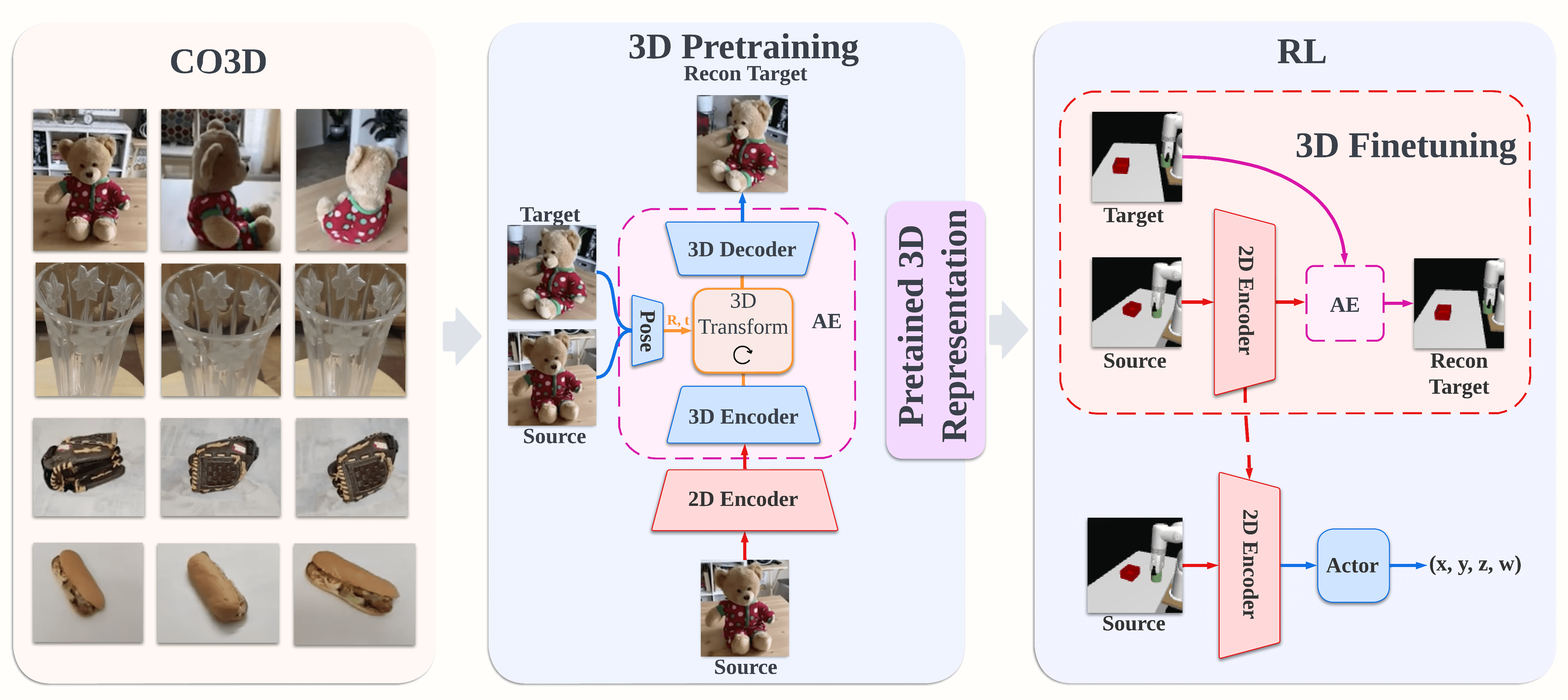}
    \caption{\textbf{Overview of our approach.} We pretrain a 3D deep voxel-based auto-encoder on the \emph{Common Objects in 3D} (CO3D) dataset. Then we train an RL policy in simulation using the learned representation as initialization and jointly finetune the representation with 3D and RL objectives on in-domain data collected by the RL agent.}
    \label{fig:overview}
\end{figure*}

In this paper, we propose a 3D representation learning framework for RL that includes both a pretraining phase using external data and a joint training phase using in-domain data collected by the RL agent. Figure \ref{fig:overview} provides an overview of our method. In the first phase, we learn a generalizable 3D representation using a repurposed \emph{video autoencoder} \cite{Lai2021VideoAS} that performs 3D deep voxel-based novel view synthesis without assuming access to ground-truth cameras. For pretraining, we leverage Common Objects in 3D (CO3D)~\cite{reizenstein21co3d} -- a large-scale object-centric 3D dataset -- to steer learning towards object-centric scene representations suitable for our downstream manipulation tasks. In the second phase, we finetune the learned representation together with policy learning on in-domain data collected by online interaction. Concretely, a 2D encoder produces a 2D feature map that is shared between the two tasks and the 3D voxel is generated upon this feature map. For the view synthesis task, we apply a random affine transformation to the voxel representation, corresponding to a change of camera pose, and task a 3D decoder with reconstructing the scene from the novel view. This encourages the network to learn the underlying scene geometry. The policy learns to predict actions from the 2D feature map, and we backpropagate gradients from both objectives to the shared encoder for in-domain finetuning. We emphasize the different views are only utilized in training and the learned model \textbf{only requires a single view for deployment}.

To validate our method, we consider a set of vision-based Meta-World \cite{yu2019meta} tasks, as well as four robotic manipulation tasks with camera feedback both in simulation and the real world as shown in Figure~\ref{fig:sim2real}. For the latter, we train policies in simulated environments, and transfer zero-shot to a real robot setup with only approximate geometric correspondence and an uncalibrated third-person RGB camera. We also demonstrate that our model is more robust to visual changes by using two variations of our real environment with different camera position, camera orientation, and lighting (bottom row in Figure~\ref{fig:sim2real}). Compared to strong baselines that pretrain representations using 2D computer vision objectives, our method demonstrates improved sample efficiency during policy learning and transfers better to the real world despite environment perturbations. In summary, our contributions are three-fold,

\begin{itemize}
    \item We propose a novel 3D representation learning framework for RL, using a view synthesis task and including a pretraining stage and a finetuning stage.
    \item Our method is evaluated on $9$ simulation tasks and achieves good sample efficiency compared to 2D representations.
    \item Our learned policy transfers zero-shot to the real world successfully in both a non-perturbed setting and a perturbed setting, showing the  robustness  of our 3D representation to visual changes.
\end{itemize}

\section{Related Work}
\label{sec:related-work}
\noindent\textbf{Representation learning for RL.} Learning good representations for vision-based RL is a well-studied problem. Prominent approaches include the use of learned dynamics models \cite{Hafner2020DreamTC, hansen2021deployment}, auxiliary objectives \cite{ hansen2021softda, Ye2021MasteringAG}, and data augmentation \cite{laskin2020reinforcement, kostrikov2020image, Hansen2021StabilizingDQ}. Recently, researchers have found that visual backbones pretrained using 2D computer vision objectives on large external datasets can produce useful features for control both in simulation and the real world \cite{Xiao2022MVP, Parisi2022TheUE, nair2022r3m}. 
Although these pretraining methods that use a \emph{frozen} visual representation have shown initial success, \emph{the domain gap between the RL task and the pretraining data is still non-negligible}. In this work we find that \textbf{jointly finetuning the visual backbone on in-domain data produces better representations for RL across different representations} including ours, ImageNet pretrained, and 2D self-supervised pretrained ones, which is a neglected factor in most previous works. Compared to these stronger, finetuned baselines, our method still performs significantly better, especially during sim-to-real transfer.

\noindent\textbf{Learning 3D scene representations.} Besides the aforementioned 2D-centric techniques, there are also prior efforts in learning 3D scene representations for RL, e.g. through differentiable 3D keypoints \cite{Chen2021UnsupervisedLO, jaritz2019multi}, object-centric graphs \cite{ Tung20203DOESVO, qi2021learning}, latent 3D features \cite{Burgess2019MONetUS, tung2019learning, Lai2021VideoAS}, and neural radiance fields \cite{Li20213DNS, ichnowski2021dexnerf,driess2022reinforcement}. \revise{Notably, the proposed framework in \cite{driess2022reinforcement} shares similarities with ours; however, their approach requires multi-view images (with perfect foreground segmentations) as input and only includes experiments in simplistic environments without any actual robots (the agent is simplified as a moving end-effector point), which makes real-world deployment challenging.} Our method also learns latent 3D features, but in contrast to prior work we only use \emph{\textbf{a single fixed view}} for policy inference, which makes our method both extendable and easy to deploy in the real world.

\noindent\textbf{Sim-to-real transfer.} Transferring policies learned in simulation to the real world is a hard problem for which a number of (largely orthogonal) approaches have been proposed. For example, domain randomization \cite{ Wang2020ImprovingGI, hansen2021softda} improves transfer by artificially widening the training data distribution. Alternatively, the simulation can be iteratively adjusted to match real world data \cite{Tsai2021DROIDMT, Du2021AutoTunedST}, the learned RL policy can be adapted by finetuning in the real world \cite{Julian2020NeverSL, hansen2021deployment, Kumar2021RMARM}, or zero-shot transfer can be improved by learning a better representation \cite{jangir2022look}. We also consider the problem of sim-to-real transfer from the lens of representation learning due to its generality and not requiring real world data, which often relies on human labor for collection.
\section{Background}
\label{sec:background}
\noindent\textbf{Problem definition.} We model agent and environment as a Markov Decision Process (MDP) $\mathcal{M} = \langle \mathcal{S}, \mathcal{A}, \mathcal{T}, \mathcal{R}, \gamma \rangle$, where $\mathbf{s} \in \mathcal{S}$ are states, $\mathbf{a} \in \mathcal{A}$ are actions, $\mathcal{T} \colon \mathcal{S} \times \mathcal{A} \mapsto \mathcal{S}$ is a transition function, $r \in \mathcal{R}$ are rewards, and $\gamma \in [0,1)$ is a discount factor. The agent's goal is to learn a policy $\pi$ that maximizes discounted cumulative rewards on $\mathcal{M}$. In visual RL, states $\mathbf{s}$ are generally unknown, but we can use image observations $\mathbf{o} \in \mathcal{O}$ in lieu of states, rendering it a Partially Observable MDP.

\noindent\textbf{Soft Actor-Critic} (SAC) \cite{haarnoja2018soft} is an off-policy actor-critic algorithm that learns a stochastic policy $\pi_{\theta}$ and critic $Q_{\theta}$ from an iteratively grown dataset $\mathcal{D}$ collected by interaction. Throughout, we let $\theta$ denote the combined parameter vector. The critic is learned by minimizing the Bellman error
\begin{equation}
    \mathcal{L}_{Q}(\theta;\mathcal{D}) = \mathbb{E}_{\mathbf{o}, \mathbf{a}, r, \mathbf{o}' \sim\mathcal{D}} \left[ (Q_{\theta}(f_{\theta}(\mathbf{o}), \mathbf{a}) - (r + \gamma \mathcal{V}) \right]\,,
\end{equation}
where $\mathcal{V} = Q_{\overline{\theta}}(f_{\overline{\theta}}(\mathbf{o}'), \mathbf{a}') - \alpha \log \pi_{\theta}(\mathbf{a}' | f_{\theta}(\mathbf{o}'))$ is the soft $Q$-target, $\overline{\theta}$ is a slow-moving average of $\theta$, $\alpha$ is a learnable parameter balancing entropy maximization and value function optimization, and $\mathbf{o}' \sim \mathcal{T}(\mathbf{o}, \mathbf{a}),~\mathbf{a}' \sim \pi_{\theta}(\cdot | f_{\theta}(\mathbf{o}'))$. $\pi_{\theta}$ learns to maximize an entropy-regularized expected return:
\begin{equation}
    \mathcal{L}_{\pi}(\theta;\mathcal{D}) = \mathbb{E}_{\mathbf{o} \sim \mathcal{D}} \left[ Q_{\theta}(f_{\theta}(\mathbf{o}), \mathbf{a}) - \alpha \log \pi_{\theta}(\mathbf{a} | f_{\theta}(\mathbf{o})) \right]\,,
\end{equation}
for $\mathbf{a}\sim \pi_{\theta}(f_{\theta}(\mathbf{o}))$. Actions are sampled from $\pi$ using a squashed Gaussian parameterization; see \cite{haarnoja2018soft} for further details. In this work, we focus on learning a good representation $f_{\theta}$ for SAC, but we emphasize that our framework is fully agnostic to the underlying RL algorithm.

\section{Method}
\label{sec:method}

We propose a 3D representation learning framework for RL that includes both a pretraining phase using external data and a finetuning phase using in-domain data collected by an RL agent. An overview of our approach is shown in Figure~\ref{fig:overview}.

\subsection{Object-Centric 3D Pretraining}
\label{sec:method-pretraining}

Our framework is implemented as a deep voxel-based 3D auto-encoder \cite{Lai2021VideoAS} that shares a 2D encoder with an RL policy. Given a view (image) of a 3D scene and an affine camera transformation, we task the 3D auto-encoder with reconstructing a 2D view of the scene after applying a transformation to the deep voxel representation. This task encourages the network to encode geometric scene information, which is beneficial for downstream control tasks.

\noindent\textbf{Architecture.} For brevity, we let $\theta$ denote the combined parameter vector of our network. A \emph{source} view $I_{\text{src}}$ is encoded by a 2D encoder $f_{\theta}$ to produce feature maps $Z = f_{\theta}(I_{\text{src}}),~Z \in \mathbb{R}^{C\times H\times W}$. We then reshape $Z$ into a 3D grid of dimensions {\small$(C/D) \times D \times H \times W$} and upsample the reshaped feature maps using strided transposed 3D convolutions $g_{\theta}$ to obtain our final deep voxel representation {\small$V = g_{\theta}(Z) = g_{\theta}(f_{\theta}(I_{\text{src}}))$}. Now, let $I_{\text{tgt}}$ denote a \emph{target} view (used as reconstruction target) of the same scene as $I_{\text{src}}$. To obtain a camera transformation between $I_{\text{src}}$ and $I_{\text{tgt}}$ for our 3D reconstruction task, we learn an additional \emph{PoseNet} $\mathcal{F}_{\text{pose}}$ that estimates the rotation between two views. This is necessary because common datasets do not have access to ground-truth cameras. $\mathcal{F}_{\text{pose}}$ takes the concatenation $[I_{\text{src}},I_{\text{tgt}}]$ as input and predicts relative rotation parameterized by Euler angles $[\alpha, \beta, \gamma]^{\top}$ (from which we trivially obtain rotation matrix $R$) as well as translation $t=[x,y,z]^\top$, \emph{i.e.}, $\mathcal{F}_{\text{pose}}(I_{\text{src}}, I_{\text{tgt}})\in \mathbb{R}^6$. We transform $V$ by $R,t$ and obtain a warped grid $\hat{V} = T_{R,t}(V)$, and predict the target view $I_{\text{tgt}}$ from $\hat{V}$ with a 3D decoder $h_{\theta}$.  The 3D network thus predicts $I_{\text{tgt}}$ from $I_{\text{src}}$ as $\hat{I}_{\text{tgt}} = h_{\theta}( T_{R,t}(g_{\theta}(f_{\theta}(I_{\text{src}})) ) )$, where $R,t$ are obtained from $\mathcal{F}_{\text{pose}}(I_{\text{src}}, I_{\text{tgt}})$. \revise{Intuitively, the source view $I_\text{src}$ is used to generate a 3D representation, and the target view $I_\text{tgt}$ is used as reconstruction target to supervise the 3D representation.}

\noindent\textbf{Objective.} To optimize the 3D auto-encoder and associated PoseNet, we adopt a $\ell_1$-norm reconstruction loss
\begin{equation}
\label{eq:reconstruction loss}
\mathcal{L}_{\text {recon }}\left(\hat{I}_{\text{tgt}}, I_{\text{tgt}}\right)=\lambda_{L 1}\left\|\hat{I}_{tgt}-I_{\text{tgt}}\right\|_{1}\,.
\end{equation}

\noindent\textbf{Training.} We implement the 2D encoder $f_{\theta}$ as an ImageNet-initialized ResNet18 and let the 3D encoder/decoder have relatively fewer parameters, such that the majority of trainable parameters are shared with the RL policy during finetuning. To steer learning of the encoder towards object-centric scene representations, we pretrain our network on 20 object categories from \emph{Common Objects in 3D} (CO3D)~\cite{reizenstein21co3d}, a large-scale object-centric 3D dataset. CO3D contains videos that rotate around objects and we only use raw frames. We emphasize that the video for pretraining is not limited to static scenes.

\begin{figure*}[t]
    \centering
    \includegraphics[width=0.65\textwidth]{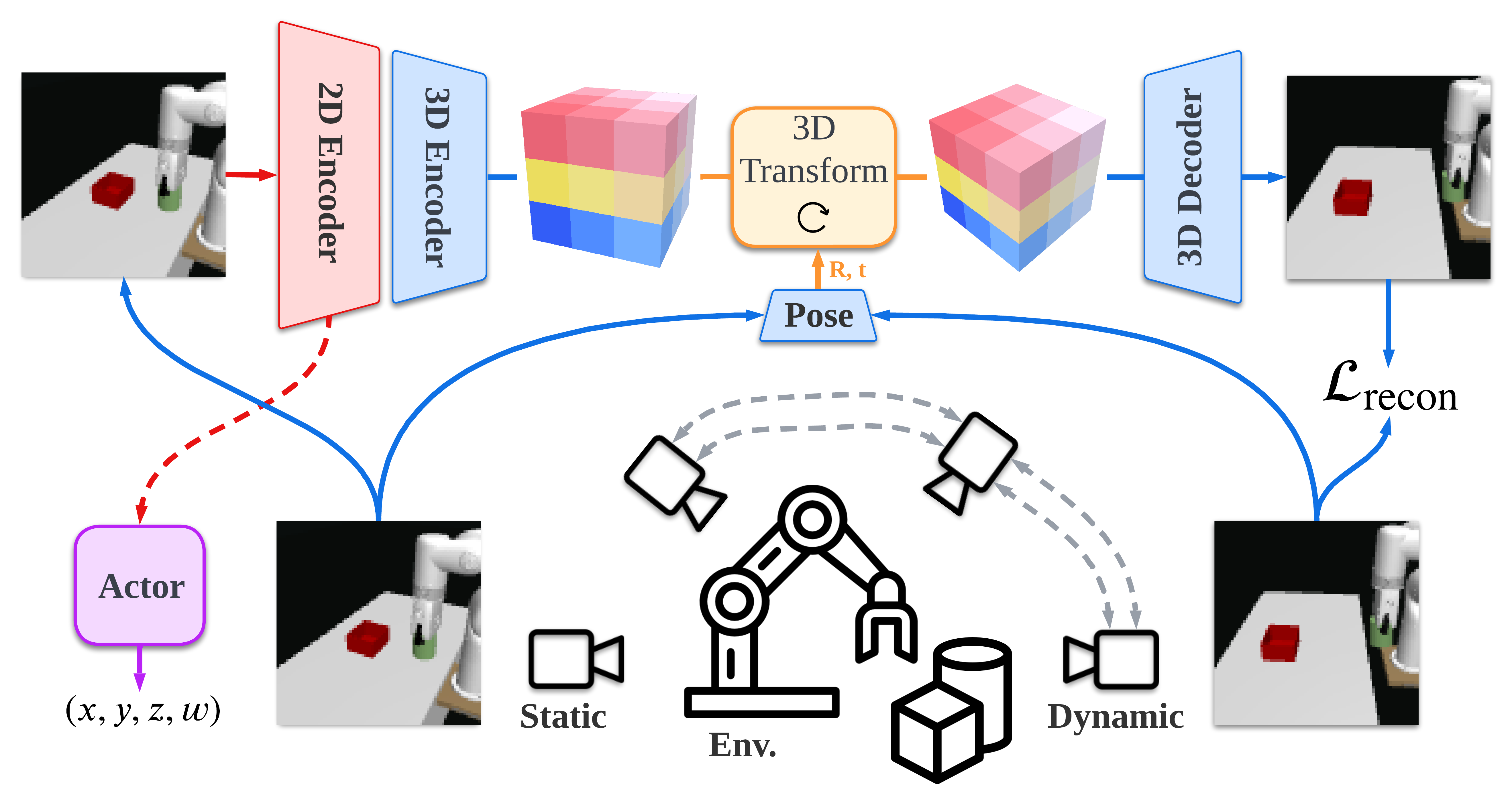}
    \caption{\textbf{In-domain joint training of 3D and RL.} A static view is used as input to both 3D and RL and is encoded using a shared 2D encoder. The 3D autoencoder takes 2D features as input and reconstructs observations from a dynamic view.}
    \vspace{-0.1in}
    \label{fig:finetuning}
\end{figure*}

\vspace{-0.1in}
\subsection{In-Domain Joint Training of 3D and RL}
\label{sec:method-finetuning}
 
After the pretraining phase, we use the learned representation as initialization for training an RL policy, while we continue to jointly optimize the 3D objective together with RL using in-domain data collected by the RL agent. Specifically, we learn a policy network $\pi_{\theta} \colon \mathbb{R}^{C\times H\times W} \mapsto \mathcal{A}$ that takes feature maps $Z = f_{\theta}(I)$ from the pretrained 2D encoder $f_{\theta}$ as input (where $I$ is an image observation) and outputs a continuous action. The motivation for our joint finetuning phase is two-fold: \emph{(1)} finetuning with the 3D objective improves the 3D representation on in-domain data, and \emph{(2)} finetuning with the RL objective improves feature extraction relevant for the task at hand. Figure \ref{fig:finetuning} provides an overview of our joint training.

\noindent\textbf{Optimizing 3D.} 
Since our proposed 3D task requires at least two views of a scene, we design a static camera and another dynamic camera. Let $I_{\text{src}}$ denote the image from the static (source) view and $I_{\text{tgt}}$ denote the image from the dynamic (target) view, respectively. The 3D task is then to reconstruct $I_{\text{tgt}}$ from $I_{\text{src}}$. We move the dynamic camera positioned with angle $\phi_d$ in a circular manner around the scene within an angle $\phi$ of the static camera, thus $\phi_d\in[0,\phi]$.

\noindent\textbf{Optimizing RL.} We train the RL agent by online interaction with a simulation environment, and store observed transitions in a replay buffer for joint optimization together with the 3D objective. 
To mitigate catastrophical forgetting in the 3D representation due to changes in the data distribution, we optimize the 3D network using a smaller learning rate than for RL. 
Formally, let $\lambda_{\text{ft}}$ denote the finetuning scale, let $\text{lr}_{\text{3D}}$ denote the learning rate for the 3D task, and let $\text{lr}_{\text{RL}}$ denote the learning  rate for RL. We then have $\text{lr}_{\text{3D}}=\lambda_{\text{ft}} \times \text{lr}_{\text{RL}}$.

\section{Experiments}
\label{sec:experiments}

We validate our method on a set of precision-based robotic manipulation tasks from visual inputs. We report success rates over a set of pre-defined goal and object locations both in simulation and in the real world.

\noindent\textbf{Robot setup} is shown in Figure \ref{fig:sim2real} \emph{(right)}. We use an xArm robot equipped with a gripper, and observations are captured by a static third-person camera. The agent operates from $84\times 84$ RGB camera observations, as well as the robot state including end-effector position and gripper aperture. 
We do \emph{not} calibrate the camera. To estimate the robustness of representations, we consider \emph{two} variants of our real-world setup of varying likeness to the simulation -- we refer to these as \emph{perturbed} and \emph{non-perturbed} environments.

\noindent\textbf{Baselines.} We implement our method and all baselines using Soft Actor-Critic (SAC; \cite{haarnoja2018soft}) as the backbone RL algorithm and use the same hyperparameters whenever applicable.  Concretely, we consider the following baselines: \emph{(i)} training an image-based SAC with a 4-layer ConvNet encoder from \textbf{Scratch}; \emph{(ii)} replacing the encoder with a ResNet18 backbone pretrained by \textbf{ImageNet} classification; and \emph{(iii)} a ResNet18 pretrained on ImageNet using the self-supervised \textbf{MoCo} \cite{chen2020improved} objective; \revise{\emph{(iv)} \textbf{CURL} \cite{laskin2020curl}, a strong visual RL method that leverages data augmentation and contrastive learning}. All methods use $\pm4$ random shift \cite{kostrikov2020image} and color jitter as data augmentation during RL, \revise{except for CURL that uses random crop augmentation as originally proposed.}

\noindent\textbf{Tasks.} We experiment with \textbf{5} image-based tasks from Meta-World, as well as \textbf{4} manipulation tasks both in simulation and on physical hardware. We consider the following tasks in our sim-to-real experiments: \emph{(1)} \textbf{reach} ($\mathcal{A} \in \mathbb{R}^{3}$), where the agent needs to position the gripper at the red goal, \emph{(2)} \textbf{push} ($\mathcal{A} \in \mathbb{R}^{2}$), where the agent needs to push a green cube to the red goal, \emph{(3)} \textbf{peg in box} ($\mathcal{A} \in \mathbb{R}^{3}$), where the agent needs to place a green peg inside a red box, and \emph{(4)} \textbf{lift} ($\mathcal{A} \in \mathbb{R}^{4}$), where the agent needs to grasp and lift a green cube into the air. A trial is considered successful only when the goal is reached (\emph{e.g.}, the peg is fully inside the box) within a fixed time limit of 20s (50 decision steps). We conduct an \emph{extensive} set of real-world trials using 5 model seeds per method per task and evaluate each seed over 10 trials (5 for reach) on a set of predefined configurations for a total of \textbf{1300} trials: \textbf{700} trials for the setup close to the simulated environments and \textbf{600} trials for the perturbed real world setup; see Figure \ref{fig:sim2real} \emph{(left)} for the two setups.

  \begin{figure*}[t]
    \centering
    \vspace{0.1in}
    \includegraphics[width=1.0\textwidth]{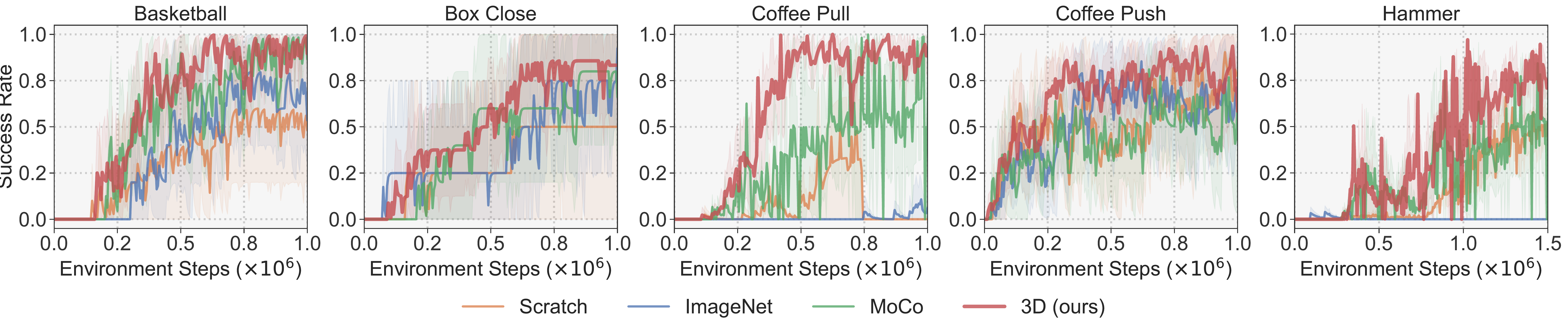}
    \includegraphics[width=0.8\textwidth]{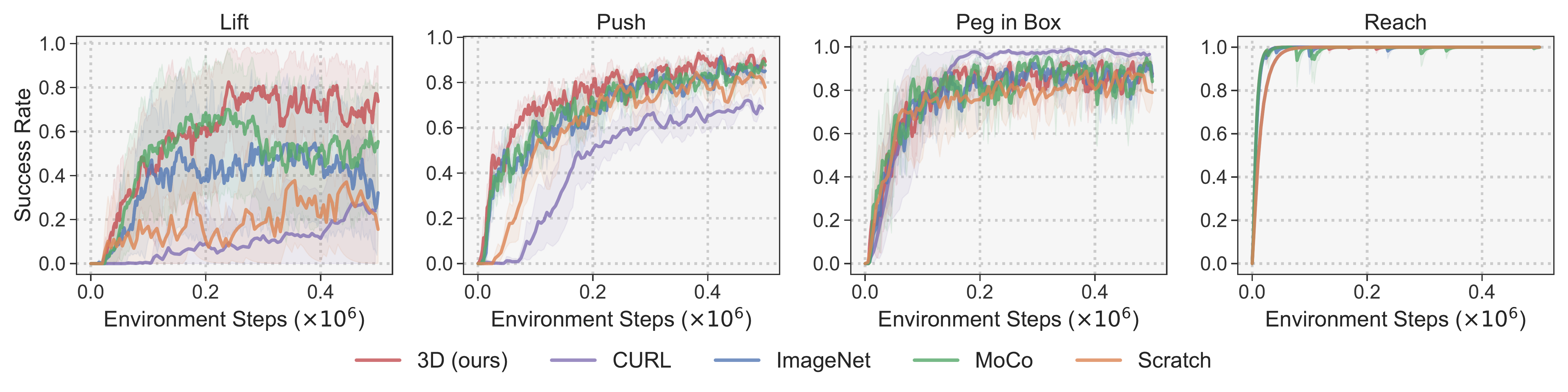}
     \vspace{-0.1in}
    \caption{\textbf{Learning curves (\emph{Meta-World and xArm})}. Success rate of our method and baselines on {\textit{five}} diverse image-based Meta-World tasks and four simulated xArm manipulation tasks. Mean of 5 seeds, shaded areas are $95\%$ CIs. Our method achieves non-trivial success rates faster than other methods.}
    \label{fig:meta-world-success-rate}
    \vspace{-0.1in}
\end{figure*}

\subsection{Sample-Efficiency}
\label{sec:results-sample-efficiency}

We train for 500k environment steps across all xArm tasks and train for 1m  environment steps across Meta-World tasks. Results for Meta-World tasks and xArm manipulation tasks are shown in  Figure~\ref{fig:meta-world-success-rate}, totalling $\mathbf{9}$ tasks. We summarize our findings as follows:

     \noindent\textbf{From scratch} training of SAC is generally a strong baseline, but the gap between this baseline and methods that use pretrained representations widens with increasing task difficulty. For example, the success rate of \emph{from scratch} is close to that of our method in \emph{coffee push} (Meta-World), while it fails to solve harder tasks like \emph{coffee pull}.

    \noindent\textbf{MoCo vs. ImageNet pretraining.} We find that MoCo generally leads to better downstream performance than pretraining with ImageNet classification, which is consistent with observations made in prior work \cite{Parisi2022TheUE}, while the performance gap is relatively small for most tasks.
    we observe MoCo to be  better on \emph{basketball} (Meta-World) and \emph{lift} (xArm) which both involve precise object manipulation. This finding suggests that self-supervised pretraining might produce better initializations for in-domain finetuning in precision-based control tasks.

     \noindent\textbf{3D vs. 2D representations.} Our proposed method that uses a self-supervised 3D representation outperforms from-scratch training, pretrained 2D representations (MoCo), \revise{and 2D representations with a SSL auxiliary objective (CURL)} across most tasks. Notably, our method enjoys large performance gains on challenging tasks such as \emph{coffee pull} (Meta-World), \emph{hammer} (Meta-World), and \emph{lift} (xArm) that require spatial understanding.

\subsection{Sim-to-Real Transfer}
\label{sec:results-sim2real}

We evaluate policies trained in simulation on physical hardware following the previously outlined evaluation procedure.  For the \emph{lift} task, we additionally report the grasping success rate in real.  Results  are shown in Table~\ref{tab:xarm-results}. We observe a drop in success rates across the board when transferring learned policies to the real world relative to their simulation performance. However, the gap between simulation and real performances is generally lower for our 3D method than for baselines. For example, our method achieves a $\mathbf{46}\%$ success rate on \emph{lift} (vs. $64\%$ in sim), whereas MoCo -- the second-best method in sim -- achieves only $20\%$ success rate (vs. $51\%$ in sim). While baseline performances differ in simulation, we do not find any single 2D method to consistently transfer better than the others.
We thus attribute the sizable difference in transfer results between our method and the baselines to the learned 3D representation.

\vspace{-0.1in}

\begin{table*}[t]
    \centering
     \vspace{0.1in}
     \caption{\textbf{Robotic manipulation results (\emph{xArm}).} Success rate (in $\%$) of our method and baselines. \emph{(left)} results in simulation. \emph{(right)} results when transferred zero-shot to physical hardware.  We report mean and std. err. across 5 model seeds for all evaluations. Initial configurations are randomized.}
    \label{tab:xarm-results}
    \vspace{-0.075in}
    \resizebox{0.496\textwidth}{!}{
        \begin{tabular}{lcccc}
            \toprule
            \texttt{Sim}     & Scratch & ImageNet & MoCo & 3D (\emph{ours}) \\ \midrule
            Reach                   & $\mathbf{100\scriptstyle{\pm0}}$ & $\mathbf{100\scriptstyle{\pm0}}$ & $\mathbf{100\scriptstyle{\pm0}}$ & $\mathbf{100\scriptstyle{\pm0}}$ \vspace{0.015in} \\ 
            
            Push                    & $65\scriptstyle{\pm16}$ & $74\scriptstyle{\pm15}$ & $74\scriptstyle{\pm14}$ & $\mathbf{80\scriptstyle{\pm14}}$ \vspace{0.015in} \\

            Peg in Box              & $77\scriptstyle{\pm22}$ & $\mathbf{82\scriptstyle{\pm 18}}$ & $\mathbf{82\scriptstyle{\pm17}}$ & $\mathbf{82\scriptstyle{\pm 17}}$ \vspace{0.015in} \\
            
            Grasp & $-$ &  $-$ &  $-$ &  $-$ \vspace{0.015in} \\ 
            
            Lift                    & $20\scriptstyle{\pm34}$ & $40\scriptstyle{\pm40}$ & $51\scriptstyle{\pm40}$ & $\mathbf{64\scriptstyle{\pm32}}$ \vspace{0.015in} \\ \bottomrule
        \end{tabular}
    }
    \resizebox{0.495\textwidth}{!}{
        \begin{tabular}{lcccc}
            \toprule
            \texttt{Real}     & Scratch & ImageNet & MoCo & 3D (\emph{ours}) \\ \midrule
            Reach                   & $84\scriptstyle{\pm12}$ & $\mathbf{96\scriptstyle{\pm4}}$ & $80\scriptstyle{\pm11}$ & $\mathbf{96\scriptstyle{\pm4}}$ \vspace{0.015in} \\ 
            Push                    & $2\scriptstyle{\pm2}$ & $22\scriptstyle{\pm10}$ & $22\scriptstyle{\pm7}$ & $\mathbf{48\scriptstyle{\pm9}}$ \vspace{0.015in} \\ 
            Peg in Box              & $40\scriptstyle{\pm14}$ & $62\scriptstyle{\pm20}$ & $50\scriptstyle{\pm15}$ & $\mathbf{76\scriptstyle{\pm19}}$ \vspace{0.015in} \\
            Grasp                   & $44\scriptstyle{\pm14}$ & $20\scriptstyle{\pm10}$ & $38\scriptstyle{\pm10}$ & $\mathbf{62\scriptstyle{\pm14}}$ \vspace{0.015in} \\
            Lift                    & $30\scriptstyle{\pm15}$ & $2\scriptstyle{\pm2}$ & $20\scriptstyle{\pm5}$ & $\mathbf{46\scriptstyle{\pm19}}$ \vspace{0.015in} \\ \bottomrule
        \end{tabular}
    }
\end{table*}
\begin{table*}[t]
    \centering
    \caption{\textbf{Robotic manipulation results evaluated in perturbed environments (\emph{xArm}).} Success rate (in $\%$) of our method and baselines. \emph{(left)} results in \emph{perturbed} (\texttt{P}) simulation environments. \emph{(right)} results when transferred zero-shot to perturbed real environments. We report mean and std. err. across 5 model seeds for all evaluations. Initial configurations are randomized.}
    \label{tab:xarm-results perturbed}
    \vspace{-0.075in}
    \resizebox{0.495\textwidth}{!}{
        \begin{tabular}{lcccc}
            \toprule
            \texttt{Sim(P)}    & Scratch & ImageNet & MoCo & 3D (\emph{ours}) \\ \midrule
             Reach                   & $76\scriptstyle{\pm10}$ & ${96\scriptstyle{\pm8}}$ & $86\scriptstyle{\pm14}$ & $\mathbf{96\scriptstyle{\pm5}}$ \vspace{0.015in} \\  
             
            Push                    & $12\scriptstyle{\pm7}$ & $12\scriptstyle{\pm10}$ & $14\scriptstyle{\pm14}$ & $\mathbf{24\scriptstyle{\pm21}}$ \vspace{0.015in} \\ 
            Peg in Box              & $20\scriptstyle{\pm20}$ & $22\scriptstyle{\pm13}$ & $24\scriptstyle{\pm7}$ & $\mathbf{34\scriptstyle{\pm20}}$ \vspace{0.015in} \\
            Grasp                   & $-$ & $-$ & $-$ & $-$ \vspace{0.015in} \\
            Lift                    & $0\scriptstyle{\pm0}$ & $10\scriptstyle{\pm15}$ & $10\scriptstyle{\pm10}$ & $\mathbf{16\scriptstyle{\pm8}}$ \vspace{0.015in} \\ \bottomrule
        \end{tabular}
    }
     \resizebox{0.495\textwidth}{!}{
        \begin{tabular}{lcccc}
            \toprule
            \texttt{Real(P)}     & Scratch & ImageNet & MoCo & 3D (\emph{ours}) \\ \midrule
            Reach                   & ${26\scriptstyle{\pm12}}$ & ${48\scriptstyle{\pm12}}$ & ${27\scriptstyle{\pm12}}$ & $\mathbf{60\scriptstyle{\pm12}}$ \vspace{0.015in} \\ 
            
            Push              & ${10\scriptstyle{\pm7}}$ & ${10\scriptstyle{\pm 7}}$ & ${0\scriptstyle{\pm0}}$ & $\mathbf{33\scriptstyle{\pm 17}}$ \vspace{0.015in} \\

            Peg in Box                  & ${18\scriptstyle{\pm11}}$ & ${28\scriptstyle{\pm14}}$ & ${20\scriptstyle{\pm6}}$ & $\mathbf{52\scriptstyle{\pm14}}$ \vspace{0.015in} \\

             Grasp                   & ${25\scriptstyle{\pm11}}$ & ${10\scriptstyle{\pm10}}$ & ${35\scriptstyle{\pm19}}$ & $\mathbf{40\scriptstyle{\pm15}}$ \vspace{0.015in} \\
            
            Lift                    & ${10\scriptstyle{\pm10}}$ & ${0\scriptstyle{\pm0}}$ & ${10\scriptstyle{\pm10}}$ & $\mathbf{25\scriptstyle{\pm11}}$ \vspace{0.015in} \\ \bottomrule
        \end{tabular}
    }
\end{table*}

\subsection{Robustness}

We provide a more challenging evaluation in both simulation and the real world by adding further perturbations to the environments. Perturbations added to the simulation includes camera position and orientation, lighting, and the texture of objects and robot. Perturbation added to the real world include camera position and orientation, as well as lighting. \revise{We refer to the latter setting as \emph{Real (Perturb)} and visualize it in Figure \ref{fig:sim2real}. Results are shown in Table \ref{tab:xarm-results perturbed}}. We observe a drop in success rates across all methods due to the perturbation, while the perturbation effects are alleviated in our method. For example, our method still achieves $\mathbf{95}\%$ success rate in perturbed simulation and $\mathbf{60}\%$ success rate in perturbed real on \emph{reach} whereas MoCo achieves only $86\%$ in sim and $27\%$ in real respectively. We also find that for 2D baselines there is no single method that outperforms others consistently. For example, ImageNet pretraining leads to better generalization on \emph{reach} while MoCo performs well on \emph{lift}. The experiments demonstrate that our 3D visual representation is more robust to distribution shifts in the observation space.

\subsection{Ablations}

\begin{figure}[t]
    \centering
    \includegraphics[width=0.5\textwidth]{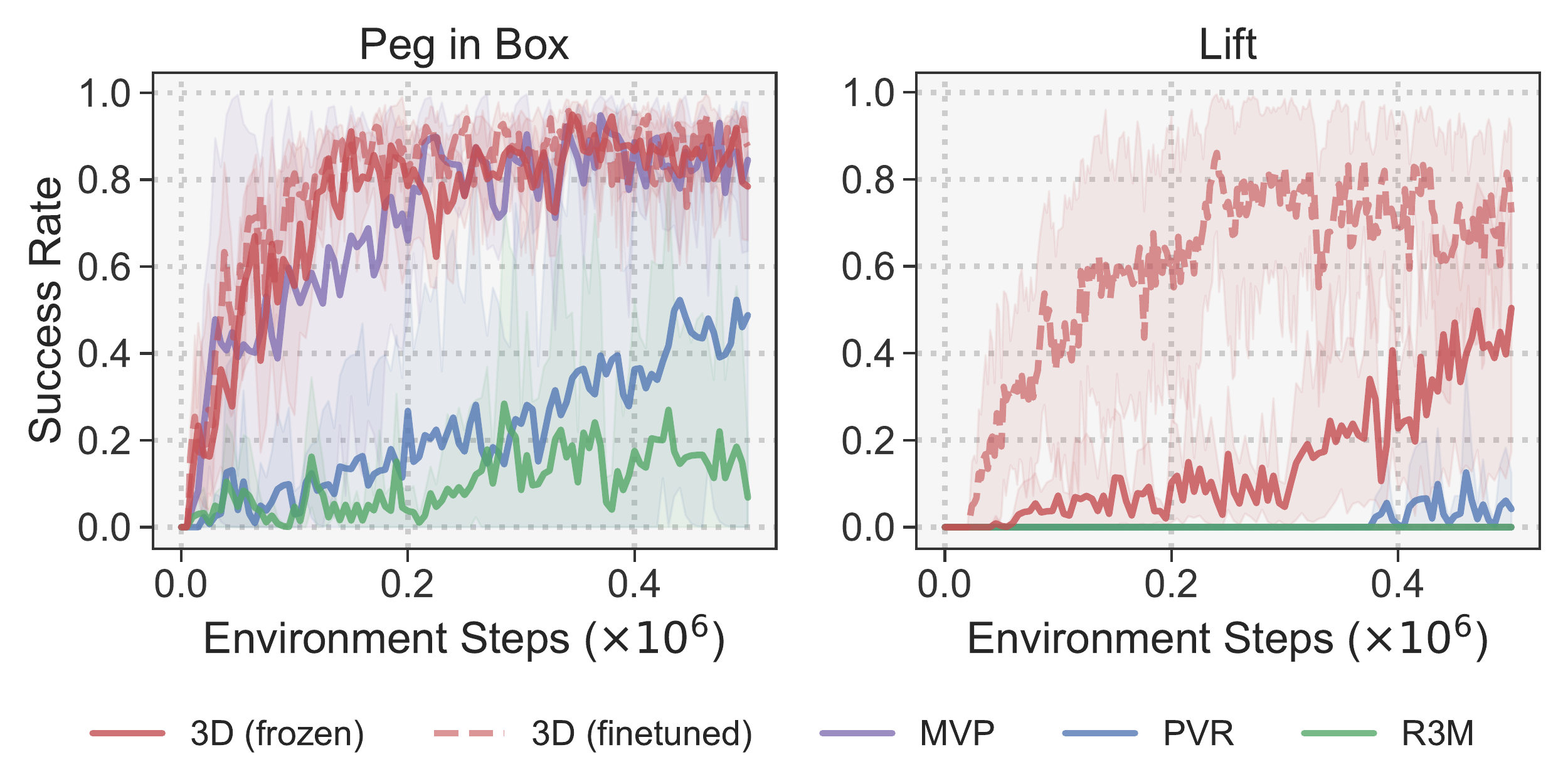}
   \caption{\small{\textbf{Success rate of different frozen visual representations.} We compare our 3D visual representation with MVP \cite{Xiao2022MVP}, PVR \cite{Parisi2022TheUE}, and R3M \cite{nair2022r3m} on \emph{peg in box} and \emph{lift}.}}
    \label{fig:quality of visual representation}
     \vspace{-0.1in}
\end{figure}

\noindent \textbf{Frozen 3D visual representation.} We compare our \textit{frozen} 3D visual representation with the following pretrain methods for motor control:
\emph{(i)} \textbf{MVP} \cite{Xiao2022MVP} which provides a  pretrained vision transformer using masked auto-encoder on a joint Human-Object Interaction dataset; \emph{(ii)} \textbf{PVR} \cite{Parisi2022TheUE} which utilizes a ResNet50 pretrained with MoCo on ImageNet; and \emph{(iii)} \textbf{R3M} \cite{nair2022r3m} which pretrains a ResNet50 with time contrastive learning and video-language alignment on the Ego4D human video dataset. The results are shown in Figure \ref{fig:quality of visual representation}.  We directly apply the public pretrained encoders provided by these works and all the methods are equipped with the same RL backbone. Our encoder uses fewer parameters (\textbf{11.47m}) than MVP (21.67m), PVR (23.51m), and R3M (23.51m). On \emph{peg in box} our method could achieve high success rates in 500k steps and is comparable to MVP, while another two baselines learn much slower. On a more challenging task \emph{lift}, only our method achieves meaningful accuracy and all other three methods have not yet. Thus our visual representation with much fewer parameters is very competitive to recent methods. In addition, we compare between the frozen 3D representation and the finetuned 3D representation and find that unfreezing the representation could give a more promising result. It is not surprising, but recent works \cite{Xiao2022MVP,nair2022r3m,Parisi2022TheUE} only focus on the frozen visual representation, which might neglect the power of end-to-end policy learning \cite{levine2016end}.

\begin{table*}[t]
    \centering
    \vspace{0.1in}
    \caption{\textbf{Quantitative evaluation of  novel view synthesis} in sim (\texttt{S}) and real (\texttt{R}) on \emph{peg in box}.  \emph{(left two)} Different $\lambda_{\text{ft}}$ for $\phi=30^\circ$. \emph{(right two)} Different dynamic camera angle $\phi_d$ when $\phi=30^\circ$ and $\lambda_{\text{ft}}=0.01$. Our method generalizes well to out-of-distribution camera angles.}
    \label{tab:view synthesis}
    \resizebox{0.24\textwidth}{!}{
        \begin{tabular}{lcc}
            \toprule
             
             ${\lambda_{\text{ft}} }$ (\texttt{S})   &  SSIM$\uparrow$ &  PSNR$\uparrow$ \\ \midrule
            
            ${0.00}$ & $8.69$ & $0.22$\vspace{0.015in} \\ 
            
            ${0.01}$ &  $11.34$ & $\mathbf{0.37}$\vspace{0.015in} \\ 
            
            ${0.10}$ & $11.75$ & $0.35$\vspace{0.015in} \\ 
             
            ${1.00}$ & $\mathbf{11.93}$ & $\mathbf{0.37}$ \vspace{0.015in} \\

           \bottomrule
        \end{tabular}
    }
    \resizebox{0.24\textwidth}{!}{
        \begin{tabular}{lcc}
            \toprule
            ${\lambda_{\text{ft}} }$ (\texttt{R})    &  SSIM$\uparrow$ &  PSNR$\uparrow$ \\ \midrule
            
            ${0.00}$ & $8.28$ & $0.28$\vspace{0.015in} \\ 
            
            ${0.01}$ &  $10.49$ & ${0.31}$\vspace{0.015in} \\ 
            
            ${0.10}$ & $11.20$ & $0.34$\vspace{0.015in} \\ 
             
            ${1.00}$ & $\mathbf{11.29}$ & $\mathbf{0.38}$ \vspace{0.015in} \\

           \bottomrule
        \end{tabular}
    }
    \resizebox{0.24\textwidth}{!}{
         \begin{tabular}{lcc}
           \toprule
            ${\phi_d}$ (\texttt{S})    &  SSIM$\uparrow$ &  PSNR$\uparrow$ \\ \midrule
            
        $15$ & $\mathbf{12.26}$ & $\mathbf{0.42}$\vspace{0.015in} \\
        $30$ & $11.34$ & $0.37$\vspace{0.015in} \\
        $45$ & $11.68$ & $0.37$\vspace{0.015in} \\
        $60$ & $10.13$ & $0.33$\vspace{0.015in}\\

           \bottomrule
        \end{tabular}
    }
    \resizebox{0.24\textwidth}{!}{
         \begin{tabular}{lcc}
           \toprule
             ${\phi_d}$ (\texttt{R})   &  SSIM$\uparrow$ &  PSNR$\uparrow$ \\ \midrule
            
        $15$ & $\mathbf{11.78}$ & $\mathbf{0.36}$\vspace{0.015in} \\
        $30$ & $10.49$ & $0.31$\vspace{0.015in} \\
        $45$ & $9.94$ & $0.24$\vspace{0.015in} \\
        $60$ & $8.72$ & $0.20$\vspace{0.015in}\\
           \bottomrule
        \end{tabular}
    }
\end{table*}
\begin{figure*}[t]
    \centering
    \begin{minipage}{1.00\textwidth}
        \centering
        \begin{minipage}{0.020\textwidth}
            \centering
            \rotatebox{90}{
                \centering
                ~~~~\textbf{Push}~~~~~~~~~~~~\textbf{Peg in Box}~~~~~~~~~~~
            }
        \end{minipage}\hspace{0.01in}
        \begin{minipage}{0.15\textwidth}
            \centering
            \textbf{Static (real)\\Input}\vspace{0.03in}\\
            \includegraphics[width=\textwidth]{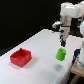}\vspace{0.005in}\\
             \includegraphics[width=\textwidth]{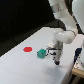}\vspace{0.005in}\\
        \end{minipage}\hspace{0.01in}
        \begin{minipage}{0.15\textwidth}
            \centering
            \textbf{Static (sim)}\vspace{0.16in}\\
            \includegraphics[width=\textwidth]{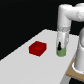}\vspace{0.005in}\\
            \includegraphics[width=\textwidth]{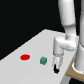}\vspace{0.01in}\\

        \end{minipage}\hspace{0.01in}
        \begin{minipage}{0.15\textwidth}
            \centering
            \textbf{Dynamic (sim) $15^\circ$}\vspace{0.03in}\\
            \includegraphics[width=\textwidth]{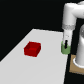}\vspace{0.005in}\\
           \includegraphics[width=\textwidth]{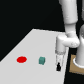}\vspace{0.005in}\\
        \end{minipage}\hspace{0.01in}
        \begin{minipage}{0.15\textwidth}
            \centering
            \textbf{Synthesis (real) $15^\circ$}\vspace{0.03in}\\
            \includegraphics[width=\textwidth]{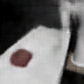}\vspace{0.005in}\\
            \includegraphics[width=\textwidth]{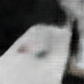}\vspace{0.005in}\\
        \end{minipage}
          \begin{minipage}{0.15\textwidth}
            \centering
            \textbf{Dynamic (sim) $30^\circ$ }\vspace{0.03in}\\
            \includegraphics[width=\textwidth]{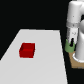}\vspace{0.005in}\\
            \includegraphics[width=\textwidth]{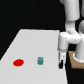}\vspace{0.005in}\\
        \end{minipage}
         \begin{minipage}{0.15\textwidth}
            \centering
            \textbf{Synthesis (real) $30^\circ$}\vspace{0.03in}\\
            \includegraphics[width=\textwidth]{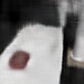}\vspace{0.005in}\\
            \includegraphics[width=\textwidth]{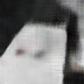}\vspace{0.005in}\\
        \end{minipage}
    \end{minipage}\hspace{0.05in}
    \caption{\textbf{Novel view synthesis in real.} We use images in the real world to generate the deep voxel and use the static view and the dynamic in the simulation to predict the transformation and then reconstruct the novel view. We display the reconstruction results for $\phi_d=15^\circ,30^\circ$ in two tasks.}
    \label{fig: view synthesis real}
    \vspace{-0.1in}
\end{figure*}
\begin{figure}[t]
\centering
\begin{subfigure}{0.155\textwidth}
\includegraphics[width=1.0\textwidth]{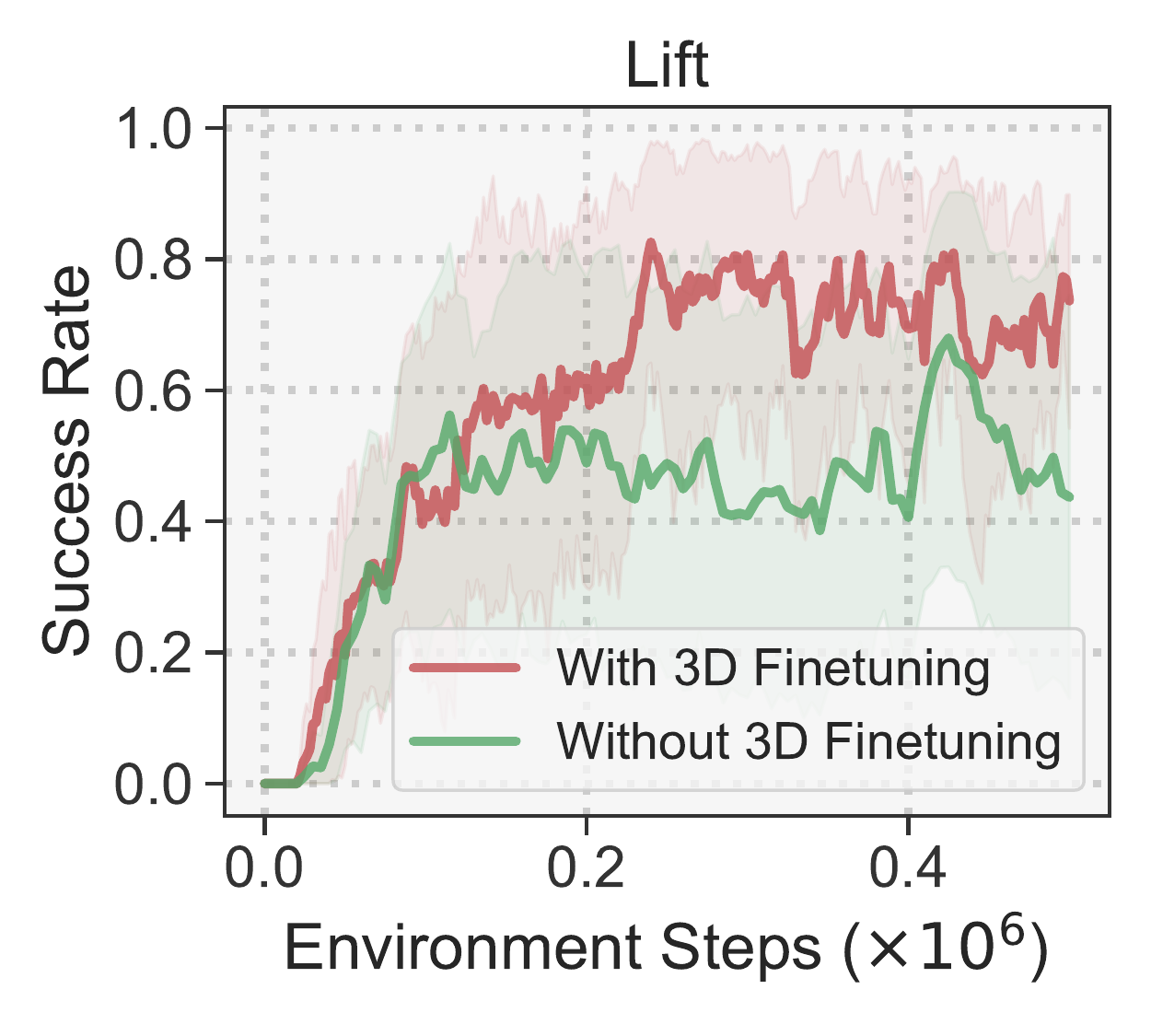}
\end{subfigure}
\centering
\begin{subfigure}{0.155\textwidth}
\includegraphics[width=1.0\textwidth]{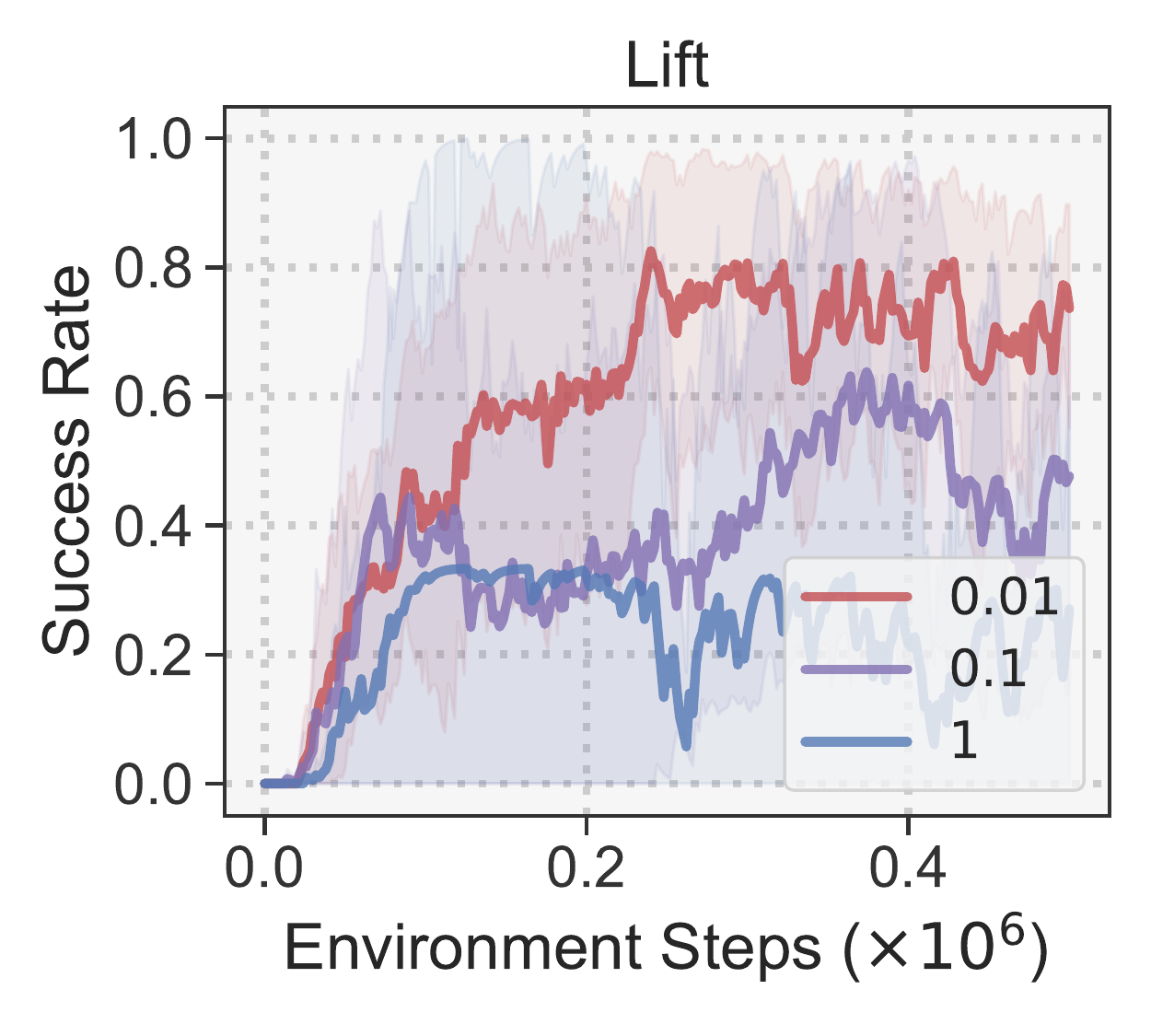}
\end{subfigure}
\centering
\begin{subfigure}{0.155\textwidth}
  \includegraphics[width=1.0\textwidth]{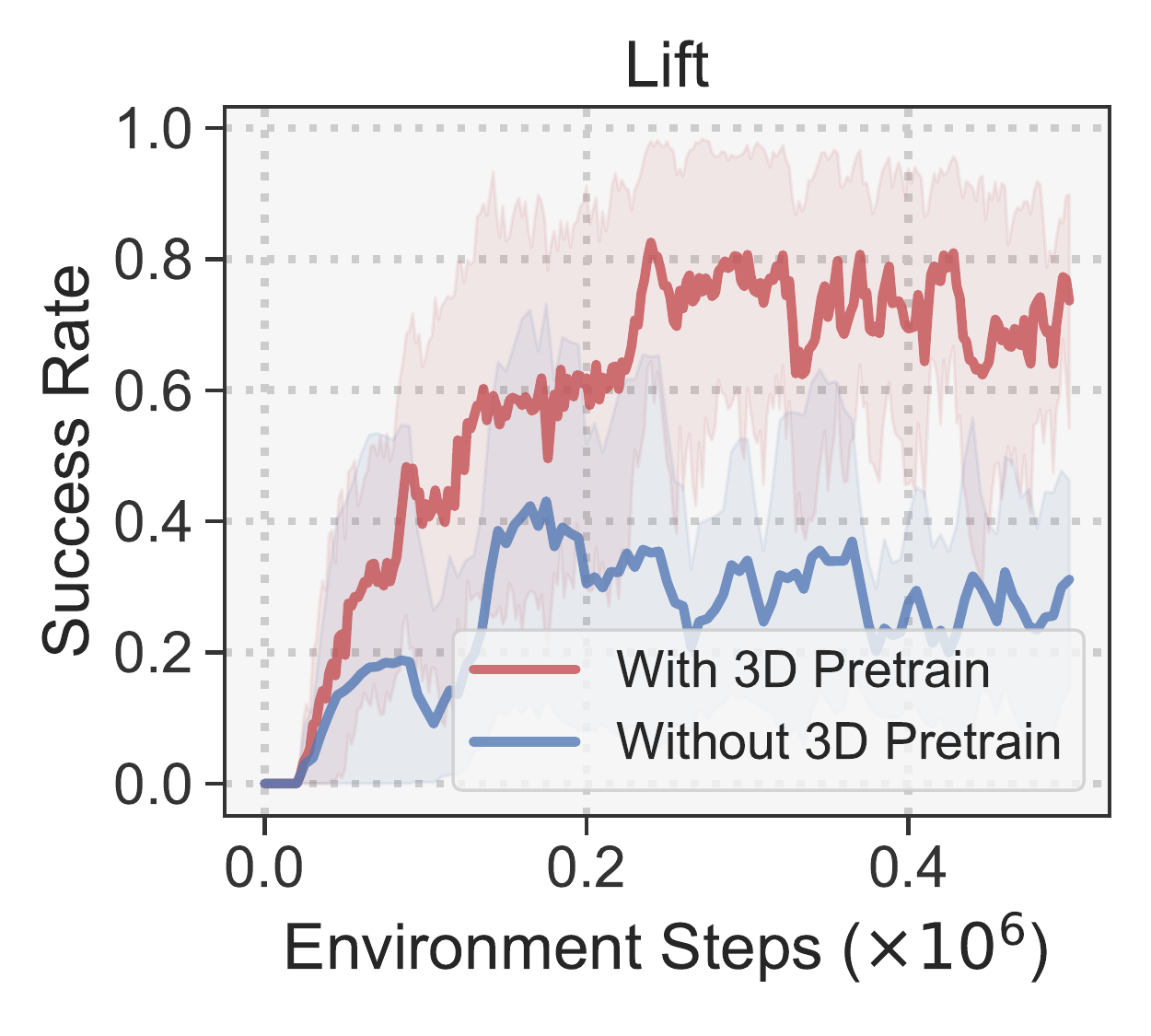}
\end{subfigure}
\centering

 \caption{\emph{(left)} Success rate of our 3D method with and without 3D finetuning on \emph{lift}. \emph{(mid)} Success rate of different $\text{lr}_{\text{ft}}$ on \emph{lift}, where $\text{lr}_{\text{ft}}$  is the finetuning scale for 3D auxiliary task such that the learning rate would be $\text{lr}\times \text{lr}_{\text{ft}}$. \emph{(right)} Success rate of our 3D method with and without 3D pretraining.
 }\label{fig: ablations}  
 \vspace{-0.1in}
\end{figure}
 \begin{table}[htbp]
    \centering
    \caption{\revise{\textbf{Quality of pose estimation for the \emph{peg in box} task.} The table shows the root mean squared error (RMSE) of camera pose estimation given varying dynamic camera angles $\phi_d$ and varying finetuning coefficients $\lambda_{\text{ft}}$. We only train with $\phi_d=30^\circ$ and evaluate both in-domain (ID) and out-of-domain (OOD) performance. We observe that finetuning leads to smaller errors.}}
    \label{tab:pose estimation}
   \resizebox{0.43\textwidth}{!}{
         \begin{tabular}{lcccc}
          \toprule
          \multicolumn{5}{c}{Pose Estimation Error (RMSE$\downarrow$)} \\\midrule
            ${\phi_d}$/$\lambda_{\text{ft}}$   &  Pretrain &  $0.01$ & $0.10$ & $1.00$ \\ \midrule
            
        $15$ (ID) & $0.041$ & $0.041$ & $\mathbf{0.040}$ & $0.046$  \\
        
        $30$ (ID) & $0.066$ & $0.059$ & $\mathbf{0.033}$ & $0.041$ \\
        
        $45$ (OOD) & $0.120$ & $0.064$ & $0.046$ & $\mathbf{0.044}$  \\
        
        $60$ (OOD) & $0.174$ & $\mathbf{0.130}$ & $0.132$ & $0.133$ \\
        
        avg & $0.100$ & $0.073$ & $\mathbf{0.063}$ & $0.066$ \\

          \bottomrule
        \end{tabular}
    }
    \vspace{-0.2in}
\end{table}

\noindent\textbf{Finetuning with the 3D objective.} We demonstrate the necessity of finetuning the 3D visual representation with in-domain data and our 3D objective, specifically the reconstruction loss. Figure \ref{fig: ablations} \emph{(left)} shows that our method without 3D finetuning initially converges at a similar rate but to a lower accuracy in 500k steps. The finetuning scale for the 3D objective also matters, as shown in Figure \ref{fig: ablations} \emph{(mid)}, where a smaller scale stabilizes the learning process. Table \ref{tab:view synthesis} provides quantitative and qualitative evaluation on the view synthesis results with 3D finetuning, demonstrating that it leads to more realistic and closer-to-original images.

\noindent\textbf{3D Pretraining.} We are also curious about whether 3D pretraining really helps if the visual representation has been trained with the 3D objective and the RL objective jointly. As shown in Figure \ref{fig: ablations} \emph{(right)}, it is observed that without 3D pretraining the representation achieves much lower accuracy on \emph{lift}. Compared to Figure \ref{fig: ablations} \emph{(left)}, we could also find that the lack of 3D pretraining leads to more degradation of the success rates, showing that 3D pretraining is necessary.

\noindent\textbf{Novel view synthesis in sim and real.}  We evaluate our method's 3D representation using qualitative and quantitative methods: \emph{(i)} novel view synthesis results from simulated and real observations (Figure \ref{fig: view synthesis real}) and \emph{(ii)} quantitative ablations to investigate the effect of important hyperparameters on reconstruction quality (Table \ref{tab:view synthesis}). Our method can synthesize meaningful reconstructions using real camera observations, even without prior exposure to our robot setup. For quantitative evaluation, we use Structural Similarity (SSIM) and Peak Signal-to-Noise Ratio (PSNR) metrics to evaluate different finetuning rates ($\lambda_{\text{ft}}$) and camera angles ($\phi_{d}$) at $\phi=30^\circ$. Results show that a reduced learning rate does not harm reconstruction, and our method trained at $\phi=30^\circ$ generalizes well to out-of-distribution angles, up to $60^\circ$.

\noindent\textbf{Camera pose estimation with PoseNet.}
Our PoseNet estimates the relative pose between two frames. We quantitatively evaluate such pose estimation results in the following. 
For a whole trajectory, we estimate the relative pose between the dynamic camera and the static camera for each timestep.
To align predicted camera pose trajectories with the ground truth \cite{Lai2021VideoAS}, we apply Umeyama alignment \cite{umeyama1991least} in deep voxel coordinate space. Our method is tested with in-domain and out-of-domain dynamic camera angles and various finetuning scales. Results (Table \ref{tab:pose estimation}) show that our method reduces pose estimation error compared to CO3D-only pretrained networks.  Our method could also generalize to 45 degrees with a small error equal to 0.064, nearly half of the one with only pretraining.
Larger finetuning scales generally reduce error, but even small scale finetuning can improve results over the pretrained model.

\section{Conclusion}
\label{sec:conclusion}
Our proposed 3D framework for pretraining and joint learning improves sample efficiency of reinforcement learning (RL) in simulation and successfully transfers to a real robot setup. This is, to the best of our knowledge, the first positive sim-to-real transfer result using pretrained 3D representations with RL. We find that learning 3D representations leads to significant gain in real robot performance and our representation is much more robust to the visual environment changes in the real world. We also compare to settings on RL with frozen features and show frozen 3D representation consistently outperforms state-of-the-art methods with frozen 2D representations. A possible limitation of our work is the requirement of multi-view inputs \emph{during training}. While these are easy to obtain in simulation, it could be difficult to learn a multi-view representation in the real world. Exploring single-view training with a 3D-aware objective could therefore be an interesting direction for future research.

\ifCLASSOPTIONcaptionsoff
  \newpage
\fi

\bibliographystyle{IEEEtran}
\bibliography{main}

\end{document}